# Extensible Knowledge Representation: the Case of Description Reasoners

**Alex Borgida**                                                BORGIDA@CS.RUTGERS.EDU
*Dept. of Computer Science*
*Rutgers University*
*New Brunswick, NJ 08904 USA*

## Abstract

This paper offers an approach to extensible knowledge representation and reasoning for the Description Logic family of formalisms. The approach is based on the notion of adding new concept constructors, and includes a heuristic methodology for specifying the desired extensions, as well as a modularized software architecture that supports implementing extensions. The architecture detailed here falls in the normalize-compared paradigm, and supports both intentional reasoning (subsumption) involving concepts, and extensional reasoning involving individuals after incremental updates to the knowledge base.

The resulting approach can be used to extend the reasoner with specialized notions that are motivated by specific problems or application areas, such as reasoning about dates, plans, etc. In addition, it provides an opportunity to implement constructors that are not currently yet sufficiently well understood theoretically, but are needed in practice. Also, for constructors that are provably hard to reason with (e.g., ones whose presence would lead to undecidability), it allows the implementation of incomplete reasoners where the incompleteness is tailored to be acceptable for the application at hand.

## 1. Introduction and Motivation

Description Logics (DLs) are a family of object-centered formalisms for representing knowledge about and reasoning with individuals grouped into classes (here called *concepts*) and related by binary relations (here called *roles*). Descriptions usually have a term-like notation that uses concept constructors and identifiers to build definitions of more complex concepts from simpler ones. For example, the description in Figure 1 is supposed to capture the noun phrase "*A collection of objects that are books and that are written by two or more authors, who are all Venusians*".

```
and(
    BOOK
    at-least(2,authoredBy)
    all(authoredBy, VENUSIAN) )
```

Figure 1: An example description.

This is accomplished by using the concept constructor **and** to conjoin terms that represent component notions:

**BOOK:** objects that are books — a concept declared elsewhere, probably as a primitive;

**at-least(2,authoredBy):** objects that are related to at least 2 other objects by the **authoredBy** role; the concept constructor is **at-least** here;

**all(authoredBy, VENUSIAN):** objects that are related by the **authoredBy** role only to objects that are in the concept **VENUSIAN**; the concept constructor here is **all**.

The concept **VENUSIAN** might itself be defined as a being whose address includes the planet value Venus: **and(BEING, all(address, fills(planet, Venus)))**. (Note that descriptions can be nested.) A more precise introduction to DLs is presented in Section 2.

Description logics reason both about intensional notions such as concepts, and about extensional aspects having to do with individuals that can be ascribed descriptions and participate in specific relationships. DLs have found a variety of applications in areas such as data management (Borgida, 1995), software engineering (Devanbu & Jones, 1997), configuration management (Wright et al., 1993), as well as general AI.

A particular description language is characterized, among others, by the choice of term constructors in it, and the significant features of DLs are clear, precise semantics and terminating reasoning algorithms for tasks such as determining whether a concept is coherent, or whether it is more general than another one.

A common difficulty faced both by designers and users of knowledge-base management systems (KBMSs) based on DLs (or any other logic, for that matter) is that many applications need to keep information about specialized kinds of data, including strings, dates, pictures, sequences of values of various kinds, etc., and it is practically impossible to anticipate all of these as part of language design.

A related problem is that if reasoning is to be complete and efficient, or even decidable, then what can be expressed in the language must be limited. For example, if in Figure 1 we also wanted to say that the authors speak the language in which the book is written, there is a concept constructor already discussed in the literature, called **subset-of**, with which one could express such a constraint as **subset-of( [writtenIn] [authoredBy,speaks] )**. Unfortunately, reasoning in a language that supports constructors **and**, **all** and **subset-of** is known to be undecidable (Schmidt-Schauss, 1989). This means that the selection of concept constructors in the language is a matter of very careful consideration for the system designer, who is faced with several choices: (a) Select a particular subset of constructors for which a sound and complete reasoning algorithm is known; if the language is sufficiently limited, this procedure is guaranteed to be "fast" (e.g., in polynomial time); otherwise, its worst case complexity is non-polynomial, though in practice the algorithm may behave well. Living with "limited languages" is however not always easy (Doyle & Patil, 1991). (b) Choose a larger set of constructors (possibly one which is even undecidable), and implement an incomplete reasoner; this however requires having to explain to the user which inferences will or will not be made. Both alternatives have different shortcomings, but, from our viewpoint, a significant common problem is that in all cases it is the designer of the DL system who makes these decisions *ahead of time*, leaving the user to sort out the consequences later on.

We propose to attack the above problems by starting with a relatively small, kernel language and system, and then providing facilities to *extend it by adding new concept constructors*. Such extensions are not to be undertaken lightly, since they provide opportunities for errors. Some of the extensions will be standard DL constructors for which complete reasoning can be implemented. Such extensions could be readily available in a library[1] . Other extensions will involve standard DL constructors which are known to be hard to reason with, and for which only an incomplete reasoning algorithm is provided, because, for example, the problem is undecidable or because the only algorithms currently known involve combinatorial search. In this second case, consultation with the user may help determine which inferences the implementation should make. For example, in classic, for the sake of a clearer semantics the system designers chose to draw no inferences about an individual based on its presence in a concept (Borgida & Patel-Schneider, 1994), although *some* of these inferences are not hard to implement. Finally, new concept constructors can be added for notions that are either domain specific (e,g,. plans, time), or which are of great practical utility but whose interaction with the full spectrum of DL constructors is not yet fully understood theoretically; such notions include keys/unique identifiers (Borgida & Weddell, 1997) and part-whole relationships (Padgham & Lambrix, 1994).

To support this, we develop a well-modularized architecture for DL-KBMSs that are implemented using the "normalize-compare" approach (see Section 2.4); this architecture expects a set of procedures to be filled in for each new concept constructor extending the original language. In addition, we propose methodological heuristics for specifying what these procedures need to do. Note that our goal is to obtain close to the same efficiency as would have been offered by a custom-built DL reasoner.

Of course, the approach presented here is not a panacea for knowledge representation and reasoning. First of all, to the extent that normalize-compare algorithms are unable to reason in a *complete* manner with DL constructors involving incomplete knowledge such as disjunction, the present system is also likely to suffer the same deficiencies. Second, the present work has not yet addressed DL notions such as role constructors, recursive concepts, and general constraints (to be described later). Finally, there are many other notions in knowledge representation, such as the full spectrum of epistemic and other non-monotonic reasoning, abduction, case-based reasoning, etc., which are likely to require a thorough overhaul of the entire reasoning architecture, and hence are likely not to be accommodated properly by the present approach.

The outline of the rest of the paper is as follows: Section 2 provides an introduction to DLs, their syntax and semantic description, and the services provided by KBMSs concerning reasoning with concepts, especially the "subsumption" relationship. Section 3 introduces the architecture of the proposed protodl approach to DL reasoning, provides an overview of the methodology for extending it, and illustrates it with constructors involving dates; it terminates by discussing successes and limitations of the proposed protodl approach to extension, an its relationship to one particular other approach that is directly relevant.

In Sections 4 and 5 we repeat the above process, but considering this time reasoning about individuals, and focusing on how to support efficiently *incremental* updates to the knowledge base.

We conclude by discussing relevant related work and summarizing the contributions and limitations of the protodl approach.

---

1. It is important to clarify from the beginning that the addition of constructors is often not a simple incremental process: some constructors may behave well independently, but cause problems when brought into the same language.

## 2. Description Logics: An Introduction

DLs are used to describe situations using various kinds of *individuals*, related by *roles*, and grouped into *concepts*. Roles that are restricted to be (partial) functions are distinguished, and are called *attributes*.

In this section we present the syntax and semantics of DLs, as well as outlining the interaction with a typical DL-based KBMS, and some implementation strategies.

### 2.1 Syntax and Semantics

As illustrated in Figure 1, DLs provide a compositional and structured language for talking about these kinds of things. Composite concepts are obtained according to the syntax presented in Table 1, which includes the concept constructors mentioned in this paper. The meta-symbols have the following referents: $CN$ is a concept name, $p$ is an atomic role (including an attribute), $f$ is an attribute, $C$ and $D$ are general concepts, $b$ is an individual, while $n$ is an integer; subscripts may occasionally be added to the above.

$$RoleChain ::= [p_1, \ \ldots \ , p_n]$$
$$AttributeChain ::= [f_1, \ \ldots \ , f_n]$$
$$C ::= \text{THING} \ | \ \text{NOTHING} \ | \ CN$$
$$| \ \textbf{and}( \ C_1, \ \ldots, \ C_n \ ) \ | \ \textbf{at-least}(n,p) \ | \ \textbf{at-most}(n,p)$$
$$| \ \textbf{all}(p,C) \ | \ \textbf{some}(p,C) \ | \ \textbf{fills}(p,b) \ | \ \textbf{one-of}(b_1, \ \ldots \ b_n)$$
$$| \ \textbf{same-as}(AttributeChain_1, \ AttributeChain_2)$$
$$| \ \textbf{subset-of}(RoleChain_1, \ RoleChain_2)$$

Table 1: Syntax of Concept Constructors Used.

To give meaning to the above syntactic terms, one can give descriptions a denotational semantics using an interpretation $\mathcal{I}=(\Delta^{\mathcal{I}}, \cdot^{\mathcal{I}})$. $\mathcal{I}$ starts by assigning to each concept name a subset of the domain $\Delta^{\mathcal{I}}$, to each role a subset of $\Delta^{\mathcal{I}} \times \Delta^{\mathcal{I}}$, to each attribute an element of $\Delta^{\mathcal{I}} \times \Delta^{\mathcal{I}}$ restricted to be functional, and to each individual some element of $\Delta^{\mathcal{I}}$. The interpretation is then extended to composite terms as follows. First, role chains (resp. attribute chains) are interpreted as mappings resulting from relation composition:

$$[p_1, \cdots, p_n]^{\mathcal{I}} = \{x \mapsto S_x \ | \ S_x = \{y \mid \exists z_1, ..., z_{n+1}. \ z_1 = x \wedge z_{n+1} = y \wedge \bigwedge_{i=1}^{n} (z_i, z_{i+1}) \in p_i^{\mathcal{I}}\}\}$$

Table 2 then presents the interpretation of complex terms using the interpretation of their components.

Alternatively, by noting that concepts are like unary predicates, and roles are like binary predicates, we can offer translation schemes from descriptions to Predicate Calculus. For example, **and(AMERICAN,OLD)** corresponds to the formula $\texttt{AMERICAN}(\gamma) \wedge \texttt{OLD}(\gamma)$, where $\gamma$ is a free variable, while **all(authoredBy,VENUSIAN)** corresponds to $\forall y.\texttt{authoredBy}(\gamma, y) \Rightarrow \texttt{VENUSIAN}(y)$.

The connection between these two kinds of specifications is that interpretations can be applied to both predicate calculus formulas as well as concepts (Baader, 1996; Borgida,

| TERM | INTERPRETATION |
|------|----------------|
| THING | $\Delta^{\mathcal{I}}$ |
| NOTHING | $\emptyset$ |
| **and(** $C_1, \ldots, C_n$ **)** | $C_1^{\mathcal{I}} \cap \ldots \cap C_n^{\mathcal{I}}$ |
| **at-least**$(n, p)$ | $\{\ d \in \Delta^{\mathcal{I}} \mid |p^{\mathcal{I}}(d)| \geq n\ \}$ |
| **at-most**$(n, p)$ | $\{\ d \in \Delta^{\mathcal{I}} \mid |p^{\mathcal{I}}(d)| \leq n\ \}$ |
| **all**$(p, C)$ | $\{\ d \in \Delta^{\mathcal{I}} \mid p^{\mathcal{I}}(d) \subseteq C^{\mathcal{I}}\ \}$ |
| **some**$(p, C)$ | $\{\ d \in \Delta^{\mathcal{I}} \mid p^{\mathcal{I}}(d) \cap C^{\mathcal{I}} \neq \emptyset\ \}$ |
| **fills**$(p, b)$ | $\{\ d \in \Delta^{\mathcal{I}} \mid b^{\mathcal{I}} \in p^{\mathcal{I}}(d)\ \}$ |
| **one-of**$(b_1, \ldots, b_m)$ | $\{\ b_1^{\mathcal{I}}, \ldots, b_m^{\mathcal{I}}\ \}$ |
| **same-as**$(FC_1, FC_2)$ | $\{\ d \in \Delta^{\mathcal{I}} \mid FC_1^{\mathcal{I}}(d) = FC_2^{\mathcal{I}}(d) \wedge FC_1^{\mathcal{I}}(d) \neq \emptyset\ \}$ |
| **subset-of**$(RC_1, RC_2)$ | $\{\ d \in \Delta^{\mathcal{I}} \mid RC_1^{\mathcal{I}}(d) \subseteq RC_2^{\mathcal{I}}(d)\ \}$ |

Table 2: Interpretation of DL Constructors.

1996): given a description $C$, with corresponding logical formula $\Psi_C(\gamma)$, and an interpretation $\mathcal{I}$, the denotation $C^{\mathcal{I}}$ is identical to the set of values $d \in \Delta^{\mathcal{I}}$ for which $\Psi_C(\gamma)^{\mathcal{I}}$ is true, when $\mathcal{I}$ is extended to map $\gamma$ to $d$.

Finally, we remark that many DLs also allow *role constructors*; for example, `isAuthorOf` can be defined as the converse of the `authoredBy` relation, by writing **inverse**(`authoredBy`). In this paper we do not consider role constructors (hence equating roles with role names).

## 2.2 Subsumption Reasoning with Concepts

The real significance of DLs is that one can *reason* about descriptions. Traditionally, the standard question one asks is whether one description is more general than (*subsumes*) another. For example, we would expect that the description in Figure 1 subsumes the following description, which requires in addition that the books be published in at most the four countries enumerated, and that there be at least three (rather than two) authors, who are to be married to earthlings:

```
and(
    BOOK
    all(publishedIn, one-of(Usa,France,Germany,Italy))
    at-least(3,authoredBy)
    all(authoredBy, and( VENUSIAN,
                         all(marriedTo, TERRESTRIAL)))
    )
```

Formally, subsumption between concepts $C$ and $D$, written as $C \implies D$, holds iff $C^{\mathcal{I}} \subseteq D^{\mathcal{I}}$ for all interpretations $\mathcal{I}$. In addition, we say that a description $C$ is *incoherent* iff its denotation $C^{\mathcal{I}}$ is the empty set in all interpretations $\mathcal{I}$. Note that this is the case iff $C \implies$ NOTHING.

The semantics of concept constructors, and especially the deductions performed by a particular system can also be specified using proof theory. A proof-theoretic specification in the "natural semantics" style (Borgida, 1992a), would indicate, among others, that greater bounds on the number of role fillers for role $p$ lead to more specialized concepts:

$$\frac{m \geq n}{\text{at-least}(m, p) \Longrightarrow \text{at-least}(n, p)}$$

This rule can be paraphrased as *"IF one can prove $m \geq n$ THEN one can write as the next line of a proof that $\text{at-least}(m, p) \Longrightarrow \text{at-least}(n, p)$."* Such a rule allows us to prove that `at-least(2,authoredBy)` $\Longrightarrow$ `at-least(1,authoredBy)`.

As usual, it is possible (and desirable) to demonstrate that the rules of inference are sound and complete with respect to the denotational semantics. Rules of inference are useful in characterizing reasoners that are incomplete with respect to the standard denotational semantics, by describing either the inferences performed, or the ones missing. Also, as we shall argue below, rules of inference can form a good starting point for developing implementations.

### 2.3 DL Concept KBMS

A concept knowledge base CKB (also known as a T-box) records constraints on concept names, including *definitions* (such as the concept `VENUSIAN` mentioned in our examples) and *necessary conditions for primitive concepts* (e.g., a `BOOK` would be required to have at least one author). In some DLs it is possible to state general subsumption constraints between arbitrary descriptions, but this will not be permitted in this paper.

Formally, CKB is a tuple $(\mathcal{R}, \mathcal{F}, \mathcal{C}, \mathcal{O}, \mathcal{N}, \mathcal{D})$ where $\mathcal{R}, \mathcal{F}, \mathcal{C}, \mathcal{O}$ are respectively the sets of role, attribute, concept and individual object identifiers declared. Concept names are either *primitive/atomic concept names*, $PN$, which have associated a necessary condition $PN \doteq C$ in the set $\mathcal{N}$ of necessary conditions; or *defined concept names*, $DN$, which have associated a definition $DN \doteq C$ in the set $\mathcal{D}$ of definitions. An interpretation $\mathcal{I}$ is a model of $PN \doteq C$ iff $PN^{\mathcal{I}} \subseteq C^{\mathcal{I}}$, and is a model of $DN \doteq C$ iff $DN^{\mathcal{I}} = C^{\mathcal{I}}$. An interpretation is a model of CKB if it is a model of all conditions in $\mathcal{N}$ and $\mathcal{D}$.

$C$ is said to subsume $D$ in the presence of a knowledge base CKB (written CKB $\models C \Longrightarrow D$), iff $C^{\mathcal{I}} \subseteq D^{\mathcal{I}}$ for all models $\mathcal{I}$ of CKB.

In this paper we are restricting ourselves to non-recursive knowledge bases, where definitions and necessary conditions are given at the same time as the name is declared, and they can only involve previously declared identifiers.

A DL-based *knowledge base management system* supports certain update operations, which affect the CKB $(\mathcal{R}, \mathcal{F}, \mathcal{C}, \mathcal{O}, \mathcal{N}, \mathcal{D})$ as follows:

| Operation | Effect |
|---|---|
| DECLARE-PRIMITIVE-ROLE($p$) | $p$ is added to $\mathcal{R}$ |
| DECLARE-PRIMITIVE-ATTRIBUTE($f$) | $f$ is added to $\mathcal{F}$ |
| DECLARE-INDIVIDUAL($b$) | $b$ is added to $\mathcal{O}$ |
| DECLARE-PRIMITIVE-CONCEPT($PN$,$D$) | $PN$ is added to $\mathcal{C}$, and $PN \leq D$ to $\mathcal{N}$ |
| DECLARE-DEFINED-CONCEPT($CN$,$D$) | $CN$ is added to $\mathcal{C}$, and $CN \doteq D$ to $\mathcal{D}$ |

In return, we expect the KBMS to respond to inquiries, which include retrieving the declarations entered and at least the following operations:

| Question | Answer type | Response |
|---|---|---|
| ASK-SUBSUMES?$(C, D)$ | Boolean | true iff CKB $\models C \Longrightarrow D$ |
| ASK-ANCESTORS$(C)$ | SET(ConceptName) | $\{CN \in \mathcal{C} \mid CKB \models C \Longrightarrow CN\}$ |
| ASK-IS-INCOHERENT?$(C)$ | Boolean | true iff CKB $\models C \Longrightarrow$ NOTHING |

A number of other operations on concepts have been found to be useful, including (i) computing the least common subsumer of two concepts (Cohen, Borgida, & Hirsh, 1992) (useful for machine learning), (ii) matching concepts against patterns (Borgida & McGuinness, 1996) (useful in large KBs for viewing only relevant aspects of concepts), (iii) finding what *additional* information has been deduced about a concept or especially an individual given the information in the KB (see Section 4.2).

To facilitate answering the above questions, and others like them, the DL-KBMS almost always performs *concept classification*: named concepts ("classes") are organized into the so-called IsA hierarchy, finding for each class the most specific other classes that subsume it. The classification algorithm relies on the $\Longrightarrow$ relationships, treating it as a subroutine, and as such is largely DL-independent. Interesting previous work in this area has been reported by Baader et al (Baader et al., 1994).

The KBMS is also charged with a number of clerical tasks, including keeping a symbol table of the declarations, maintaining and accessing efficiently the precomputed IsA hierarchy, signaling definitions/declarations that are redundant (i.e., a concept which is equivalent to a previously defined one) or are incoherent.

DL-KBMS perform inferences on individuals, as well as concepts. The operations involved will be described in Section 5.

## 2.4 Reasoning Strategies

There are two general approaches to answering the fundamental subsumption question underlying the DL-KBMS operations.

One approach is based on theorem proving techniques specially adapted for descriptions, particularly variants of tableau techniques that determine the subsumption A $\Longrightarrow$ B by checking for the unsatisfiability of $A \land \neg B$; systems such as KRIS (Baader et al., 1994),CRACK (Bresciani, Franconi, & Tessaris, 1995), FaCT (Horrocks, 1998) and DLP (Patel-Schneider, 1998) follow this approach. Such systems have the advantage of being provably complete. Although the worst-case complexity bounds of the subsumption problems are sometimes quite high (EXPTIME complete), recent empirical evidence shows that their performance for computing subsumption on large realistic KBs is quite effective (Horrocks & Patel-Schneider, 1998).

All the implemented DL KBMS systems that have had wide distribution and use, such as BACK (von Luck et al., 1987), CLASSIC (Borgida et al., 1989), and LOOM (MacGregor, 1986), follow a so-called "normalize-compare" paradigm, where most of the reasoning work is performed in an initial "normalization" phase, whose goal is to find a normal form for concepts which explicates implicit facts, eliminates redundancies and detects incoherencies. Once this is done, when, for example, it comes to comparing two concepts for subsumption, it is often (but not always) a matter of comparing only "structurally similar" elements – ones built with the *same* constructor. For example, since the descriptions

**all(pet,**NOTHING**)** and **at-most(0,pet)** are equivalent (subsume each other), in a normal form both would appear, so that subsumption comparison with **all(pet,**DOG**)** would only require looking at **all(pet,**NOTHING**)**, while comparison with **at-most(3,pet)** would locate **at-most(0,pet)**.

The present work was carried out over a period of several years within this "normalize-compare" paradigm. The original reasons for this choice included the following features, which were only available for this paradigm: having a sizable user base; supporting additional KBMS operations, such as least common subsumer, pattern matching, explanation; reasoning with large knowledge bases of individuals, and especially doing so incrementally in an environment where one needs the "logical completion" of an individual.

We remark that the tableau technology has recently become competitive on sizable knowledge bases of concepts. The promise of provably complete and effective reasoners for expressive languages is very enticing, and therefore the study of extensibility in tableaux approaches (Baader & Hanschke, 1991; Bresciani et al., 1995) is of great interest.

## 3. Concept Reasoning in PROTODL

Our approach to extensible KBMSs is based on the idea that in a case when the current KBMS does not meet the users' representation and reasoning needs, one can extend it by adding one or more new description constructors – what would be considered logical connectives.

Although this approach is clearly restrictive, it has the advantage of being very specific and directed; it allows us to develop an implementation architecture and accompanying methodology to support the process of extension. The approach is also supported by empirical evidence from the use of the CLASSIC system in many applications: the test-defined concepts of CLASSIC have been crucial in escaping the expressive limitations of the basic language, and each such test function acts essentially as a new concept constructor[2].

### 3.1 The Modularized Implementation Architecture

As previously described, the concept reasoning services of our DL-KBMS are delivered by operations which rely on the $\Longrightarrow$ relationship, to be computed by the function `Subsumes?`. In our implementation paradigm, this function takes as arguments two *normalized concepts*. Hence we need some way to take input and convert it into normalized concepts. This means parsing the concept (function `Parse`) and normalizing it (function `Normalize`).

Because of the uniform prefix nature of DL syntax, parsing is easily performed by recursive descent with a single token look-ahead; so we associate with every concept constructor **Q**, a function **Q::Parse**.

After examining earlier implementations and applications, a key decision underlying the PROTODL implementation is that a normalized concept is viewed as the *conjunction* of a collection of component concepts, which are either concept identifiers or are normalized terms built with concept constructors **Q** other than **and**. Therefore, it is possible to view a normalized concept as a data abstraction that encapsulates the way in which the various components are stored and accessed thru a set of functions `put_Q(value)` and `get_Q()` — one for each concept constructor **Q**. For example, the **put_one-of** operation stores a set of individuals `S`, while **get_one-of** returns this set[3]. On the other hand, for the constructor **fills**, the (normalized) value stored is a set of terms of type `<Role × SET(Individual)>`, since for each role, we need to store the set of current fillers. For constructors such as **fills**, which are associated with a single role, we overload the **put** and **get** functions so they accept a separate role argument, allowing us to access each value independently (e.g., `get_fills(p:<Role>)` returns `<SET(Individual)>`). In fact, for such constructors it is more efficient to group them by the role identifier, but to simplify the exposition we shall not pursue this distinction in the rest of this section.

---

2. Note however that reasoning with test-concepts in CLASSIC was minimal: no subsumption reasoning other than identity checking was performed, and individual reasoning was limited to recognition.

As mentioned earlier, the ideal normalized concept would have the property that subsumption could be determined by comparing, using **Q::Subsumes?**, only components built with the same constructor. In other words, **Subsumes?** would have the pseudo-code

```
Subsumes?(hiNC, lowNC: <NormalizedConcept>) returns <Bool>
    ;; structural subsumption
    {For every Q(T) in getComponents(hiNC) do
        { lowComponent := get_Q(lowNC);
            if not Q::Subsumes?(T,lowComponent)
                return false; }
    return true; }
```

As we shall see in Section 3.3, this is too restrictive however, so we will pass as parameter the entire normalized lower concept, `lowNC`, rather than just the `lowComponent`. Therefore, for every constructor **Q** we expect a function

$\boxed{\texttt{Q::Subsumes?(<NormalizedQ-term>,<NormalizedConcept>)}}$ .

In order to normalize a concept, we will normalize each component separately, and then combine them. Since a normalized concept represents a conjunction, the combinator function will be named `Conjoin`. Here are then the remaining functions that are provided ahead of time by the core implementation:

```
Normalize(pC : <ParsedConcept> )    returns <NormalizedConcept>
                                    throws IncoherentExn;
    /* Start ND to be a data structure having no constraints – i.e. like THING */
     ND := copy of NormalFormTHING;
  {  /* Normalize each component and conjoin it onto ND */
        For every Q(T) in pC do {
            T_norm := Q::Normalize(T);
            Q::Conjoin(T_norm,ND); }
        /* Replace  concept ids by  normalized form  (saved by DECLARE ) */
            For every concept identifier N in pC do {
                N_norm :=  look up normalized form of N;
                Conjoin(N_norm,ND); }
```

---

3. The type of a parameter will be indicated by enclosing it in angle-brackets. Thus, we would have referred to **put_one-of**(s: `<SET(Individual)>`).

```
    /* if any Normalize or  Conjoin detects an incoherence, propagate it up,
            since the whole conjunct is incoherent. */
    }     catch IncoherentExn { throw IncoherentExn }
return ND;

Conjoin(fromNC, ontoNC: <NormalizedConcept>) throws IncoherentExn
    /* Take each component of  fromNC and conjoin it to  ontoNC;
       modifies ontoNC. */
  {For every  Q(T) in getComponents(fromNC) do
      {Q::Conjoin( T, ontoNC)
      } catch IncoherentExn {throw IncoherentExn }
  }
```

Therefore, for every constructor **Q**, we also expect corresponding functions `Q::Normalize`
and `Q::Conjoin`. Note that the software architecture treats incoherent concepts (ones
equivalent to NOTHING) in a special manner: whenever they are encountered, a special ex-
ception `IncoherentExn` is raised. Conjunction propagates these exceptions, but other con-
structors may trap them and handle them in their own way. For example, **all**`::Normalize`
accepts the restriction being incoherent, but must ensure that **at-most**(0) is also added to
the normalized concept for the same role.

In retrospect, the above three functions (`Subsumes?, Normalize` and `Conjoin`) really
represent the semantics of the constructor **and** (and hence can be appropriately called
**and**`::Subsumes?`, etc.) as well as the expansion of necessary conditions/definitions for
named concepts ("inheritance"), and parts of the meaning of the built-in concepts NOTH-
ING and THING. To make this more complete, we can add to the beginning of the `Subsumes?`
function the statement: `if equal(hiNC,NormalForm`THING`) or equal(lowNC,NormalForm `NOTH-
ING`) then return true;`

In this sense, PROTODL starts from a minimal, core language specified by the syntax

$$C ::= \text{THING} \mid \text{NOTHING} \mid CN \mid (\textbf{and } C_1, \ldots, C_n)$$

This is not to say that everyone must start from this minimalist language, which is hardly
useful. But it will have the advantage that all constructors (even the most useful ones,
like **all**, **at-least**, etc.) will be implemented as extensions, according to the same uniform
paradigm as the more esoteric extensions. This will be essential if we are to be able to
easily *modify* the implementation of standard constructors (like **all**) when called upon to
implement non-standard ones (like dates), which may interact with them.

### 3.2 An Overview of the Process of Extension.

Suppose we want to extend the system at some particular stage with a new concept con-
structor. The following is a suggested methodology for accomplishing this, illustrated with
a familiar concept constructor, **all**.

**(1)** *Determine a syntax for the extension.* If concepts will have a LISP-like syntax (as in CLASSIC and LOOM), the constructor **all** might be given the syntax (`:all` *Role Concept*). The terms following the constructor are called its *arguments*, and a version of them will eventually be stored in the internal representation of our normalized concepts. One can now implement the function **all::Parse**, which in this case would invoke `Role::parse` and `Concept::parse`.

**(2)** *Determine a semantics for the new concept constructor.*
First, this requires settling on a domain of values from which its denotation will come. This might be the set of ordinary objects with unique intrinsic identity — instances of a special class ANY-OBJECT, which is a built-in direct subclass of THING. The denotation might also be some new kind of value (e.g., triples for dates, or lists of objects), in which case it is necessary to introduce (possibly as a new concept constructor, which takes no argument) a top class for these kinds of values (e.g., ANY-DATE for dates).

It is now time to clarify the intended meaning of the new constructors. One alternative, not explored here, is to express the semantics using First Order Predicate Calculus. Another is to assign a denotation of values to concepts built with the new constructor. For **all**, this is just the usual interpretation

$$\mathbf{all}(p, C)^{\mathcal{I}} = \{\ d \in \Delta^{\mathcal{I}} \mid p^{\mathcal{I}}(d) \subseteq C^{\mathcal{I}}\ \}$$

Rather than move on to implementation right away, our experience indicates that it is easier to first describe the deductions to be performed in a more concise manner: through rules of inference. Based on empirical evidence, we see these rules of inference as being naturally grouped into several categories[4]

- *Rules dealing with just one constructor,* **Q**.

  - *Normalizing the arguments of* **Q**, if these are themselves composite objects.

$$\frac{C \equiv D}{\mathbf{all}(p, C) \equiv \mathbf{all}(p, D)}$$

  - *Reasoning only with concepts of the form* **Q***(args).*
    * When is such a concept incoherent *by itself*?

$$\mathbf{all}(p, C) \text{ is never incoherent.}$$

    * When is such a concept equivalent to the entire domain of values?

$$\mathbf{all}(p, \text{THING}) \equiv \text{ANY-OBJECT}$$

    * When does **Q**(arg1) subsume **Q**(arg2) in terms of arg1 and arg2? (The so-called structural subsumption relationship.)

$$\frac{C \Longrightarrow D}{\mathbf{all}(p, C) \Longrightarrow \mathbf{all}(p, D)}$$

---

4. In each case, we give a general description of what is desired for a new constructor **Q**, and then illustrate it with the specific constructor **all**.

∗ When does **Q**(arg) entail some description built with some other constructor?

$$\mathbf{all}(p, \text{NOTHING}) \equiv \mathbf{at\text{-}most}(0, p)$$

– *Reasoning with conjunctions of descriptions built with* **Q** *only.* This usually results in a normal-form that either combines the arguments into a single one, or keeps the arguments as a list, set or multi-set.

$$\mathbf{and}(\mathbf{all}(p, C), \mathbf{all}(p, D)) \equiv \mathbf{all}(p, \mathbf{and}(C, D))$$

• *Rules of inference dealing with combinations of several concept constructors.* These rules can be grouped into similar families as above, except that they will involve two or more kinds of constructors.

Since there is no such rule involving **all**, we offer an alternative example:

$$\mathbf{and}(\mathbf{at\text{-}most}(n, p), \mathbf{at\text{-}least}(m, p)) \equiv \text{NOTHING} \;\; if \;\; n < m$$

The augmented set of inference rules should be proven sound and complete with respect to the semantics specified earlier. As with any logic, soundness is relatively easy to show, but completeness is much harder. Royer and Quantz propose one approach (Royer & Quantz, 1992) for generating inference rules that are complete, based on translation to First Order Logic, but the technique is not easy to apply.

**(3)** *Extend the implementation.* This requires first determining an appropriate normal form for the argument of constructor **Q**, declaring a data structure to hold the information, and completing the `put_Q` and `get_Q` procedures to access them from a normalized concept. For simplicity, we will not address detailed data structure issues here; instead, we use a simple notation based on terms in the style of Prolog to represent data structures. In the case of **all**, the representation is just a pair **all**(`<Role>`,`<NormalizedConcept>`). We note that if we wanted to keep track of the original form of the concept (so it can be printed out to users as entered), this could be added as an extra field of the data structure. We emphasize that the normalized form of the description (e.g., a finite automaton) might not even resemble the original syntax (e.g., a regular expression), though in our examples this will not be the case.

Next, we need to implement the three procedures: `Q::Normalize`,`Q::Conjoin`, and `Q::Subsumes?`, whose invocation is orchestrated by the corresponding functions for the **and** constructor.

By analyzing the normalization algorithms for numerous constructors, we have arrived at a more refined understanding of the *prototypical* `Q::Conjoin` procedure. This is presented in the Appendix A as Figure 3, and uses the following other, smaller functions `Q::Universal?`, `Q::Incoherent?`, `Q::SubsumesSame?`, `Q::ConjoinToSame`, `Q::IncoherentWDifferent?`, `Q::SubsumesDifferent?`, `Q::ConjoinToDifferent`, `Q::FindOtherImplications`.
These functions correspond to the categories of inference rules we have mentioned above. When this decomposition can be applied, it provides two advantages: improved chances of correct implementation by making smaller, specialized modules; and independent reuse, as in

```
Q::Subsumes?(T:<NormalizedQ-Term>, lower:<NormalizedConcept>)
                                         returns <Bool>
   {if not Q::SubsumesSame?(T,get_Q(lower)) then return false;
    if not Q::SubsumesDifferent?(T,lower) then return false;
    return true; }
```

Of course, if a specific constructor does not need to implement some of these sub-procedures, the corresponding lines from the generic implementations are omitted, in order to avoid the cost of useless function invocations.

For our simple example, when **Q**=**all**, **all::Normalize(all(role,vr))** returns **all(role, and::Normalize(vr))**.

If the only other constructor is **at-most**, the following functions used by **all:Conjoin** and **all::Subsumes?** are needed:

- **Q::Universal?(T)**: *Is* T *equivalent to the top of the hierarchy for that set of values?*
  **all::Universal?(all(role,vr))** returns true when **vr** is **NormalForm**THING.

- **Q::SubsumesSame?(T,OldTs)**: *Is* T *implied by the* **Q**-*constructed terms already seen in oldTs?* This is basically the "structural subsumption" algorithm.
  **all::SubsumesSame?(all(role,vr1),all(role,vr2))** would return true iff **and::Subsumes?(vr1,vr2)** returns true.

- **Q::ConjoinToSame(T,OldTs)**: *Merge* T *with any preceding terms built with* **Q**.
  **all::ConjoinToSame(all(role,vr1),all(role,vr2))** returns
  **all(role,and::Conjoin(vr1,vr2))**.

- **Q::FindOtherImplications(T,This,ImplicationsToDoList)**: *If any additional constructors have to be added because of* **Q**, *put them onto the* **ImplicationsToDoList**.
  **all::FindOtherImplications** adds to **ImplicationsToDoList** the description
  **at-most(0,role)** if its first argument is **all(role,**NOTHING**)**.

Once the `Conjoin, Normalize` and `Subsumes?` functions are implemented for the new constructor **Q**, the corresponding functions for **and** need to have lines added to invoke them according to the pattern presented in the preceding section (unless the programming language is highly polymorphic).

Next, we use reasoning about dates, an extension desired in some CLASSIC applications, to give a more complex example of the methodology for building extensions of PROTODL.

## 3.3 Dates: An Example Concept-Level Extension

We imagine an application where there will be individual dates as values of attributes or even roles, and that concepts will describe collections of dates, specified in various ways (e.g., as ranges or periods).

To begin with, it is important to clarify what the *individuals* look like about which information will be kept. In the case of dates, we know that we wish to treat them as temporal points, which have associated information about the year, month and day when that date occurs. Given a date $d$, the above components will be referred to as year($d$),

month($d$), day($d$). There are two different ways of thinking of a date: as an abstract mathematical value (e.g., a triple of integers) or as an individual object with attributes for `year, month, day.` The last approach has the disadvantage that we need to develop a separate theory of identity for dates (two dates are supposed to be identical if they have the same components), although it has the advantage of allowing incomplete information about the exact occurrence of a time point. Since this feature is not desired, we will adopt the first approach, for convenience writing date individuals as 1996/7/25. In addition, we will have the usual total (reflexive) order $\preceq$ on dates, as well as two special date constants, BeginTime and EndTime, such that BeginTime $\preceq$ $d$ $\preceq$ EndTime for all dates $d$.

We are now ready to introduce concept-forming operators for dates which are useful for our application. First, when introducing a new *kind* of value, it is useful to define a top concept, which will contain all such values. In our case, let us call it ANY-DATE.

One obvious grouping of dates is by ranges: "between June 1st and August 31, 1996". We might thus propose a concept constructor **dateRange**, which takes as argument a pair of dates, denoting the ends of the range, e.g., **dateRange**(`1996/6/1,1996/8/31`). But since the base language does not support disjunction, and we want to allow descriptions such as "the summers of 1995 and 1996", we will in fact have **dateRange** take as argument a set of date pairs, as in **dateRange**(`{(1995/6/1,1995/8/31)` , `(1996/6/1,1996/8/31)}`).

Having established the syntax of the concept constructors, and then implemented **date-Range::Parse**, we present its denotational semantics:

$$\textbf{dateRange}(SD)^{\mathcal{I}} = \{\ d \mid \exists b, e.(b,e) \in SD \ \text{such that} \ b \preceq \ d \preceq e\ \}$$

Next, we look for inference rules describing the desired reasoning for **dateRange**. Following the heuristics in Section 3.2, we come up with Table 3.

To implement the inferences for **dateRange**, we must find a normal form and write functions **dateRange::Normalize**, **dateRange::Conjoin** and **dateRange::Subsumes?**, or their components.

It is useful to determine first the structural subsumption function, **dateRange::-SubsumesSame?**, since this drives the requirements for the others. In our case, the obvious representation works fine: a date is some data structure, such as a list of three values or a record with three fields; a single date-pair is a list of two values or a record with two fields; and the set of date-pairs making up the disjunction is just a list of date-pairs. It is best to encapsulate the above implementation choices using abstract data types for `Date,` `DatePair` and `DateRange`, with appropriate accessor functions and constructors. We must also extend the representation of normalized concepts to implement **put_dateRange** and **get_dateRange**.

Returning to **dateRange::SubsumesSame?**, this function now simply implements the subsumption inference rules above

```
dateRange::SubsumesSame?(high,low : <SET(Date×Date)> ): returns <Bool>
    {for every (b,e) in low
        find (b',e') in high such that b' ⪯ b ⪯ e ⪯ e';
    }
```

| Universal | $\mathbf{dateRange}(\{(\text{BeginTime},\text{EndTime})\}) \;\equiv\; \textsc{any-date}$ |
|---|---|
| Incoherent | $\mathbf{dateRange}(\{\}) \;\equiv\; \textsc{nothing}$ |
| Subsumption | $$\frac{\text{b1} \;\preceq\; \text{b2} \;\preceq\; \text{e2} \;\preceq\; \text{e1}}{\mathbf{dateRange}(\{(\text{b2,e2})\}) \implies \mathbf{dateRange}(\{(\text{b1,e1})\})}$$ $$\frac{\mathbf{dateRange}(\alpha 1) \implies \mathbf{dateRange}(\alpha 2) \quad \mathbf{dateRange}(\beta 1) \implies \mathbf{dateRange}(\beta 2)}{\mathbf{dateRange}(\alpha 1 \cup \beta 1) \implies \mathbf{dateRange}(\alpha 2 \cup \beta 2)}$$ |
| Conjunction | $\mathbf{and}(\mathbf{dateRange}(\{(\text{b2,e2})\}), \mathbf{dateRange}(\{(\text{b1,e1})\})) \;\equiv\;$ $\mathbf{dateRange}(\{\;(max(\text{b1,b2}),min(\text{e1,e2}))\})$ $$\frac{\mathbf{and}(\mathbf{dateRange}(\alpha),\mathbf{dateRange}(\beta 1)) \;\equiv\; \mathbf{dateRange}(\gamma 1) \quad \mathbf{and}(\mathbf{dateRange}(\alpha),\mathbf{dateRange}(\beta 2)) \;\equiv\; \mathbf{dateRange}(\gamma 2)}{\mathbf{and}(\;\mathbf{dateRange}(\alpha), \mathbf{dateRange}(\beta 1 \cup \beta 2)) \;\equiv\; \mathbf{dateRange}(\gamma 1 \cup \gamma 2)}$$ |
| Normalize | $$\frac{\text{e} \;\preceq\; (\text{b} + 1 \text{ day})}{\mathbf{dateRange}(\{(\text{b,e}),\alpha\}) \;\equiv\; \mathbf{dateRange}(\{\alpha\})}$$ $$\frac{\text{b1} \;\preceq\; \text{b2} \;\preceq\; (\text{e1} + 1 \text{ day}) \;\preceq\; \text{e2}}{\mathbf{dateRange}(\{(\text{b1,e1}),(\text{b2,e2})\}) \;\equiv\; \mathbf{dateRange}(\{(\text{b1,e2})\})}$$ |

Table 3: Inference rules for **dateRange**.

**dateRange::Normalize** verifies (if not already done so) that the dates b and e in each pair (b,e) are valid according to our usual calendar, and that $\text{b} \preceq \text{e}$. Any pairs not satisfying these conditions are eliminated. It also needs to merge overlapping or adjacent intervals into maximally long ones, since otherwise the above subsumption algorithm will not recognize that **dateRange({(1996/1/2,1996/1/4), (1996/1/5,1996/1/6)})** subsumes **dateRange({1996/1/2, 1996/1/6})**.

The function **dateRange::Conjoin** has the standard implementation (see Appendix A), but only functions **dateRange::Universal?**, **dateRange::Incoherent?**, **dateRange-::ConjoinToSame** and **dateRange::SubsumesSame?** need to be implemented, performing exactly the actions specified by the rules of inference.

Consider now adding another concept constructor for dates, to help represent *periodic* time, such as "every summer" or "every Christmas". A single period is just a range constraint on the possible values for the components of a date, other than `year`. So, "summer days" would be represented by [(6 . 8) (1 . 31)], while "Christmas" would be [(12 . 12) (25 . 25)]. For the sake of brevity, this new **period** constructor will take as argument only a single period (rather than a set of them, interpreted disjunctively) and we will only sketch the implementation extensions. Also not presented here, is a very useful extension to periods allowing constraints on the days of the week, such as "every Saturday to Sunday", or even "every 3rd Sunday".

| Universal | **period**([(1 . 12) (1 . 31)]) ≡ ANY-DATE |
|---|---|
| Incoherent | $\dfrac{not(\ \mathrm{bm} \le \mathrm{em}\ \ and\ \mathrm{bd} \le \mathrm{ed}\ )}{\mathbf{period}([(\mathrm{bm.em})\ (\mathrm{bd.ed})])\ \equiv\ \textsc{nothing}}$ |
| Subsumption | $\dfrac{\mathrm{bm1} \le \mathrm{bm2} \le \mathrm{em2} \le \mathrm{em1}\ \ and\ \ \mathrm{bd1} \le \mathrm{bd2} \le \mathrm{ed2} \le \mathrm{ed1}}{\mathbf{period}([(\mathrm{bm2.em2})(\mathrm{bd2.ed2})]) \Longrightarrow \mathbf{period}([(\mathrm{bm1.em1})(\mathrm{bd1.ed1})])}$ |
| Conjunction | **and**(**period**([(bm1.em1)(bd1.ed1)]),   **period**([(bm2.em2)(bd2.ed2)]))<br>≡<br>**period**([($max$(bm1,bm2) . $min$(em1,em2))<br>($max$(bd1,bd2) . $min$(ed1,ed2))]) |

Table 4: Inference rules for **period**.

The denotation of **period** terms is also quite simple:

$$\mathbf{period}([(m1.m2)(d1.d2)])^{\mathcal{I}} = \{e \mid m1 \le month(e) \le m2,\ d1 \le day(e) \le d2\}$$

The rules of inference for reasoning about **period** concepts alone, presented in Table 4 are also straightforward. The implementation of most functions follows immediately from the rules of inference.

The interesting reasoning involves the conjunction of ranges with periods. These rules, appearing in Figure 5, can be expressed most succinctly by showing how certain periods are either eliminated or retained unchanged, and then relying on the **dateRange** normalization rule, applied *in reverse*, to cut up a range into appropriate strips. These inference rules imply that we need to also implement **period::ConjoinToDifferent (p:<Period>, other:<NormalizedConcept>)**, which basically uses the period as a "cookie cutter" to create the sub-interval ranges that satisfy the period's restriction. Of course, this will be done procedurally in a manner more efficient than suggested by the rules of inference. Moreover, we must now revisit the implementation of **dateRange**, to put in a similar **dateRange::ConjoinToDifferent** function, since a concept with a range may be conjoined onto one that already has a period in it.

Finally, in the presence of both **dateRange** and **period** constructors, structural subsumption is not enough, since we want **period([(4 . 4) (1 . 31)])** to subsume **dateRange((1988/4/1, 1988/4/21))**, but not **dateRange((1990/4/1, 1992/4/1))**, and there is no finite normal form which would list all the ranges satisfying a period. So we need to also write the function **period::Subsumes Different?(p:<Period>, lower:<NormalizedConcept>)**, which checks every interval **(b,e)** in **get_dateRange(lower)** to make sure that **year(b)=year(e)** and that the month and day meet the conditions of the period.

An interesting complication arises because dates are discrete, and hence one can count the number of dates in a **dateRange**. If **dateRange** can appear as the value restriction

$$\textbf{and}(\textbf{dateRange}(\alpha \cup \beta), \textbf{period}(\delta))$$
$$\equiv$$
$$\textbf{and}(\textbf{and}(\textbf{dateRange}(\alpha), \textbf{period}(\delta)), \ \textbf{and}(\textbf{dateRange}(\beta), \textbf{period}(\delta)))$$

---

$$\mathrm{bm} \leq \mathrm{month}(\mathrm{dt1}) = \mathrm{month}(\mathrm{dt2}) \leq \mathrm{em} \ \ \textit{and}$$
$$\mathrm{bd} \leq \mathrm{day}(\mathrm{dt1}) \leq \mathrm{day}(\mathrm{dt2}) \leq \mathrm{ed}$$

$$\textbf{and}(\ \textbf{dateRange}(\{\ (\mathrm{dt1,dt2})\ \}), \ \textbf{period}([(\mathrm{bm.em}),(\mathrm{bd.ed})]) \ )$$
$$\equiv$$
$$\textbf{dateRange}(\{\ (\mathrm{dt1,dt2})\ \})$$

---

$$\mathrm{month}(\mathrm{dt1})=\mathrm{month}(\mathrm{dt2}) \ \ \textit{and}$$
$$\textit{not} \ \ (\mathrm{bm} \leq \mathrm{month}(\mathrm{dt1}) \leq \mathrm{em} \ \ \textit{and} \ \ \mathrm{bd} \leq \mathrm{day}(\mathrm{dt1}) \leq \mathrm{day}(\mathrm{dt2}) \leq \mathrm{ed} \ )$$

$$\textbf{and}(\ \textbf{dateRange}(\{(\mathrm{dt1,dt2})\}), \ \textbf{period}([(\mathrm{bm.em}),(\mathrm{bd.ed})]) \ ) \ \equiv \ \textsc{nothing}$$

Table 5: Inference rules for conjoining **period** and **dateRange**.

of general roles, as in **all(freeForMeeting, dateRange({(1995/6/1,1996/6/5)}))** attributes, there is no problem. Otherwise, we can however infer cardinality constraints on the number of fillers: **dateRange({(1995/6/1,1996/6/5)})** allows only 5 different values. In a language that supports constructor **at-most**, we would therefore have to conclude **at-most(5,freeForMeeting)**. The implementation achieves this by introducing a helper function **dateRange::countDays**, which can be applied to a normalized **dateRange** object; then, function **all::FindOtherImplications** needs to be modified, so that it invokes **dateRange::countDays** on its value restriction if it is of date type, and a normalized **at-most** restriction is posted for that role on the **ImplicationsToDoList**. Later processing of that list will remove the **at-most** constraint and conjoin it onto the concept.

## 3.4 Experience with Extensions

As part of the development of the above architecture, we have considered extending the original core with the constructors of classic (**all**, **at-least**, **at-most**, **fills**, **one-of**, integer ranges), as well as primitive concept negation, and the negation of **fills**. In these cases we have reproduced the inferences of classic and almost exactly the internal actions of the classic implementation; i.e., we perform only a few more checks despite the fact that our implementation is made up of "standard" modules for each constructor.

Two kinds of concept constructors seem difficult to add to a normalize-compare algorithm in a way that preserves completeness of reasoning *and* the architecture of the system. Disjunction would most naturally be handled if there was a disjunctive normal form, where each disjunct is purely conjunctive. This is difficult to achieve with nested disjunctions (inside **all** restrictions say). Note that we had no problem with the one-level disjunction in **dateRange**.

A second kind of construct that is difficult to add efficiently and completely is **same-as**. The reason here is that **same-as** interacts with **all** in a way that generates a potentially infinite number of **all** restrictions; therefore, the implementation of **same-as** is best combined with that of **all**, resulting in a non-tree data structure (Borgida & Patel-Schneider, 1994). A speculative way to preserve structural subsumption might be to allow **all** to apply to chains of a attributes represented by regular expressions. (In general, the complications of implementing **same-as** are the reason it does not appear in C-classic and neo-classic.)

Finally, as mentioned earlier, we do not currently cover role constructors and recursive concept constraints. Furthermore, constructors that require entirely new deductive mechanism (e.g., epistemic reasoning, defaults, forward-chaining rules) will also have difficulty being integrated properly into this framework.

On the positive side, we have considered extensions supporting strings (Borgida, Isbell, & McGuinness, 1996), and most elaborately, a reconstruction of the clasp reasoner about actions and plans (Borgida, 1992b). To summarize this "success" briefly, clasp (Devanbu & Litman, 1996) was a system built on top of classic for reasoning about actions (which were represented in propositional STRIPS-style, having concepts for pre- and post-conditions, as well as add and delete lists); plans were represented by regular expressions of actions, as illustrated below. Our goal was to apply the protodl approach to clasp, hoping to be able to reproduce the original, custom-made implementation discussed by Devanbu and Litman (Devanbu & Litman, 1996). First, we introduced a concept constructor for actions, **act**(*PreC, PostC, AddC, DeleteC*), with the expected logical properties expressed as rules of inference in the various categories (e.g., if *PreC* was incoherent, then the action was also incoherent). The implementation then followed immediately.

The more interesting problem was dealing with plans. The constructors **single**, **seq**, **loop**, and **altern** can be used to build complex plans from actions, as in **seq**( **single**(`DIAL`), **loop**(**single**(`RING`))). These plans denote *sequences of action instances* (e.g., in the above example, a dialing action followed by any number of rings). Although we provided rules of inference for plans too, it turns out that there is no normal form for regular expressions! Instead, in the implementation, `Normalize` for **seq**, **altern** and **loop** built a non-deterministic finite automaton, which then had to be made deterministic, and in which certain chains of arcs had to be removed (if the post-condition of the action on the incoming edge was inconsistent with the pre-condition of the action on the outgoing edge). Moreover, the containment algorithm for these finite automata had to take into account the fact that actions on transitions (e.g., `MOVE`) may represent generalizations of others (e.g., `MOVE-FAST`). This implementation was achieved in protodl by introducing a "hidden" concept constructor, **top-plan-exp**, which enclosed the top plan. It was then **top-plan-exp::Normalize** that made the automaton deterministic and removed some transitions, and **top-plan-exp::Subsumes?** that implemented the special subsumption algorithm. Moreover, the requirement to implement **top-plan-exp::Conjoin**, which was not present in the original clasp system, made us realize that without this, clasp plans could not appear in other concepts, because expressions like **and**(**all**(`p,PLAN1`), **all**(`p,PLAN2`)) could not be normalized. As a result we believe we reconstructed and improved the original clasp proposal, by characterizing the inferences performed through rules of inference, and by allowing plans to be first-class values, which could appear in ordinary roles.

### 3.5 Relationship to "Concrete Domain" Extensions

Although we shall address general work on extensible KR&R in the conclusion, there is one specific approach involving description logics that deserves closer scrutiny at this stage, while the details of the present work are still fresh.

Baader and Hanschke's proposal (Baader & Hanschke, 1991) for extending DLs with "concrete domains" allows concepts to consist of arbitrary predicates involving values from some domain other than ordinary objects (i.e., other than elements of $\Delta^{\mathcal{I}}$), as long as these values are fillers of *attributes* of ordinary objects. For example, suppose the concrete domain is that of dates; as above, we have predicates like *BEFORE*, corresponding to $\preceq$ ; then, if we had two date-valued attributes `arrival` and `departure`, we could define the concept `BEFORE(arrival,departure)`, denoting ordinary objects (not sets of dates!) whose attributes have appropriately related date values.

In order to keep reasoning decidable, this mechanism requires the concrete domain to be *"admissible"*: (i) there must be a predicate denoting the universe of all values in that concrete domain; (ii) the set of predicates must be closed under negation; and (iii) it must be possible to decide the satisfiability of any finite conjunction of such predicates. It is interesting to note that these requirements match in part our heuristics for new concept constructors: we also argued for the need to add a top concept to the hierarchy of new values (and its negation, the bottom of this hierarchy), and for the need to be able to compute the conjunction of descriptions.

The admissibility of a concrete domain ensures that the PROTODL approach can implement any such extension as follows: Syntactically, a domain corresponds to a concept constructor. Therefore, a concept like `BEFORE(arrival,departure)` in the domain **DATE**, would be represented in PROTODL as **DATE('BEFORE',arrival,departure)**. Then, the necessary PROTODL functions are programmed as follows: **DATE::Normalize** tests that the predication is satisfiable (otherwise signaling **IncoherentExn**), and creates a singleton list containing the predication; **DATE::Conjoin** concatenates the list of predications of its arguments, checking for the consistency of their conjunction; **DATE::Subsumes?(C,D)** creates the conjunction of all the predicates in C, and their negation in D, and returns true if the result is unsatisfiable.

Conversely, Baader and Hanschke's approach could well be applied to date ranges, since these are essentially closed under negation. And since it is offered in the context of the tableau theorem-proving technique mentioned in Section 2.4, this continues to have the advantages of elegance and complete reasoning, as long as the concrete domain reasoner is proven complete. In our case, the entire system would have to be proven complete.

We believe though that PROTODL is somewhat more general, since it allows concrete domains to be value of roles, not just attributes (see our earlier discussion of dates as values of roles). Also, there is some advantage to being able to deal with non-admissible domains in cases when negation is not absolutely needed, but its addition would cause an increase in computational complexity; for example, adding negation/complement to regular expressions makes the containment problem non-elementary (Stockmeyer, 1974).

## 4. Processing of Individuals in DLs: An Introduction

Concept descriptions, as introduced above, are intensional objects, suited for capturing generic information about a domain, such as the ontology of terms. DL-KBMSs must also manage extensional/factual information about individual objects — the so-called A-box[5].

### 4.1 Inferences Involving Individuals

Normally, one can assert information such as the fact that some object `Anni` is known to be an instance of a concept `CHILD` (written as `Anni:CHILD`), or that it has some other object, `Lego` as a filler for its `hasToys` role (written as `(Anni,Lego):hasToys`). Based on this information, the KBMS can deduce information about the individual's membership in other descriptions (written using $\longrightarrow$); for example, in this case we know `Anni` $\longrightarrow$ **at-least(1,hasToys)**. Because DL-KBMS usually do not make the closed-world assumption, it is also necessary to record when some set of fillers is complete for an individual's role. This is done using an (auto-epistemic) assertion like `Anni:`**allFillersKnown(hasToys,{Lego,Barbie})**. As a result of such an assertion, we can then deduce that `Anni` $\longrightarrow$ **at-most(2,hasToys)**. (Note that we have distinguished the three kinds of *assertions* in $\mathcal{A}$, namely $b : C$, $(b, e) : p$, $b :$ **allFillersKnown**$(p, S)$, from the corresponding judgements that can be deduced in the logic: $b \longrightarrow C$, b $\longrightarrow$ **fills**(p,e), b $\longrightarrow$ **closedFillers**(p,S).)

Information about fillers and roles being closed (i.e., all fillers being known) can also be deduced, as in the case of the KB that contains $\mathcal{A}=\{$`Lori:`**all(hasToys,one-of(Lego45))**, `Lori:`**at-least(1, hasToys)**$\}$, from which we can conclude that `Lego45` is a toy owned by Lori (written as `Lori` $\longrightarrow$ **fills(hasToys,Lego45)**) and that the complete set of fillers for `hasToys` is { `Lego45` } (written as `Lori` $\longrightarrow$ **closedFillers(hasToys,**{ `Lego45` }**)**). We note that an elegant formalization of these notions has been obtained by adding an epistemic modal constructor **K** to DLs (Donini et al., 1998). This constructor has a number of other uses, but to shorten the presentation we have not introduced it here explicitly.

Formally, we define a knowledge base KB to be a concept knowledge base CKB, extended with a set $\mathcal{A}$ of *assertions* of the form $b : C$, $(b, e) : p$ and $b :$ **allFillersKnown**$(p, S)$, where $S$ is some set of individuals. An interpretation $\mathcal{I}$ is said to be a model for $b : C$ if $b^{\mathcal{I}} \in C^{\mathcal{I}}$, a model for $(b, e) : p$ if $e^{\mathcal{I}} \in p^{\mathcal{I}}(b^{\mathcal{I}})$, and a model for $b :$ **allFillersKnown**$(p, S)$ if $(e^{\mathcal{I}} \in p^{\mathcal{I}}(b^{\mathcal{I}}) \iff e \in S)$[6].

The judgment $KB \models b \longrightarrow C$ holds iff for every model $\mathcal{I}$ of KB, $b^{\mathcal{I}} \in C^{\mathcal{I}}$; the judgement $KB \models b \longrightarrow$ **fills**$(p, e)$ holds iff for every model $\mathcal{I}$ of KB, $(b^{\mathcal{I}}, e^{\mathcal{I}}) \in p^{\mathcal{I}}$; finally, $KB \models b \longrightarrow$ **closedFillers**$(p, S)$ holds iff $KB \models b \longrightarrow$ **fills**$(p, b_i)$ for every $b_i \in S$, and for every other individual $e$ not in $S$, there is some interpretation $\mathcal{I}$ of KB such that $\not\!KB \models b \longrightarrow$ **fills**$(p, e)$ . Because our concept language may not have negation, and because we have the open world assumption, we also need to be able to talk about non-membership in a concept: $b \not\longrightarrow C$; naturally, $KB \models b \not\longrightarrow C$ iff for some model $\mathcal{I}$ of KB, $b^{\mathcal{I}} \notin C^{\mathcal{I}}$. As usual, a KB will be called *inconsistent* iff it has no models.

---

5. The most thorough theoretical investigation of individual reasoning has been presented in Andrea Schaerf's PhD thesis and derived publications (Schaerf, 1994). We note however that for some of the most expressive DLs proposed recently, individual reasoning has not yet been addressed.

6. As usual, we make the unique-name assumption, and in fact include named individuals in the domain

The specification of reasoning about individuals can again be represented by inference rules (Borgida, 1992a). For example, for the constructor **all**, we proffer the following three rules of inference, which describe the "structural membership/non-membership" rules, as well as a kind of inference called "propagation", where information about some individual $b$ results in new information being deduced about *another* individual $e$:

$$\frac{KB \vdash \text{b} \longrightarrow \textbf{closedFillers}(\text{p,S}) \quad KB \vdash b_i \longrightarrow \text{C}}{KB \vdash \text{b} \longrightarrow \textbf{all}(\text{p,C})} \quad S = \{b_1, \cdots, b_n\}, \ n \geq 0$$

$$\frac{KB \vdash \text{b} \longrightarrow \textbf{fills}(\text{p,e}) \quad KB \vdash \text{e} \not\longrightarrow \text{C}}{KB \vdash \text{b} \not\longrightarrow \textbf{all}(\text{p,C})}$$

$$\frac{KB \vdash \text{b} \longrightarrow \textbf{all}(\text{p,C}) \quad KB \vdash \text{b} \longrightarrow \textbf{fills}(\text{p,e})}{KB \vdash \text{e} \longrightarrow \text{C}}$$

Note that for an inconsistent KB, we can have $KB \vdash b \longrightarrow$ nothing, or we can deduce information from fillers (e.g., $KB \vdash \text{b} \longrightarrow$ **at-least(3,pets)**) that contradicts information asserted or deduced from descriptors (e.g., $KB \vdash \text{b} \longrightarrow$ **at-most(2,pets)**).

## 4.2 DL-KBMS Operations on Individuals

The point of living with the open-world assumption is to allow information to be accumulated incrementally, as in the case of designing some artifact (one of the most successful applications of classic).

From the functional point of view, the DL-KBMS therefore supports the following update operations for incrementally adding information about individuals:

| Operation | Effect |
|---|---|
| assert-member$(b, C)$ | $b : C$ is added to $\mathcal{A}$ |
| assert-fills$(b, p, b_1)$ | $(b, b_1) : p$ is added to $\mathcal{A}$ |
| assert-closed$(b, p)$ | b:**allFillersKnown**(p,S) is added to $\mathcal{A}$, where $S$ is the set of individuals returned in the current KB by the operation ask-for-fillers$(b, p)$, defined below. |

If as a result of the update, the KB is inconsistent, then the update is rejected and the state of the KB is supposed to remain unchanged.

At any point, the KBMS is able to respond to inquiry operations about relationships involving individuals:

| Question | Answer type | Response |
|---|---|---|
| ask-member?$(b, C)$ | Boolean | true iff $KB \models b \longrightarrow C$ |
| ask-non-member?$(b, C)$ | Boolean | true iff $KB \models b \not\longrightarrow C$ |
| ask-for-fillers$(b, p)$ | SET(Individual) | $\{ e \mid KB \models b \longrightarrow \textbf{fills}(p, e) \}$ |
| ask-closed?$(b, p)$ | Boolean | true iff for some set $S$ $KB \models b \longrightarrow \textbf{closedFillers}(p, S)$ |

As with concepts, many DL-KBMS pre-compute the $b \longrightarrow C$ judgment, for all individual and concept names, by finding the most specific named descriptions to which the individual $b$ provably belongs. Similarly, the DL-KBMS pre-computes and caches the fillers and closed information for each individual's roles. This is done in order to detect inconsistencies at the time of the update, to decrease the amortized cost in case queries are much more frequent than updates, and to support queries of the form "What additional information can you deduce about $v$?". The utility of such queries has been shown, among others, in applications involving information discovery in software development (Devanbu, 1994).

## 5. Individuals in PROTODL

Reasoning about individuals The goal of our implementation is to support efficiently the update operations, *which are incremental*, so that most of the inferences are precomputed: for each individual, we know the least classes it is an instance of, all the fillers it can have for each role, and whether the role is closed or not for that individual. The architecture below describes an extension and rationalization of the individual processing that is usually carried out in normalize-compare DL-KBMS such as CLASSIC. The novelty will be the systematic separation of the interaction between various kinds of updates and various kinds of inferences, and the concomitant use of truth-maintenance links.

### 5.1 The Basic Architecture of Individual Reasoning

In some systems, one can try to reduce individual reasoning to concept reasoning by associating some single, maximally complete description with each individual. However, this is not always possible, as shown by Schaerf (Schaerf, 1994).

Instead, our data structure for every individual includes both a concept description (to be called `Descriptor`), which may be an existing class or an unnamed complex description, and *individual role-filler information* recording the specific values assigned so far, and whether the role is closed or not. (This information is accessed with built-in functions `add_filler`, `put_closed`, `get_fillers` and `is_closed?`.)

In processing an update about some individual $b$, we must therefore resort to two kinds of reasoning: (i) subsumption/incoherence reasoning involving only the description `Descriptor`($b$) and other concepts; (ii) reasoning specific to the individual and especially its role fillers.

The former kind of reasoning is motivated by general inference rules connecting membership and subsumption (e.g., if $b \longrightarrow C$ and $C \Longrightarrow D$ then $b \longrightarrow D$). Since subsumption and its extensibility has been described in the preceding section, henceforth we concentrate on the second aspect of individual reasoning.

To understand the tasks involved, let us illustrate what kinds of reasoning need to be performed whenever any fact is asserted (or inferred) about an individual $b$

- *The KB may become inconsistent*, because of a conflict between the individual descriptor and the filler information either on $b$ or some other individual. For example, more fillers might be added to a role than permitted by an **at-most** restriction. This requires the entire update to be rejected. Note that an update may result in several individuals with inconsistent information on them. The system is only required to detect one of them, before rejecting the update.

- *The individual b may end up being re-classified.* In a monotonic system, such as the present one, this means that the individual may now belong to some more specialized class(es) in the hierarchy.

- *New information about the roles of individual b can be inferred.* For example, learning that a role is closed allows us to now count its fillers and hence obtain an exact upper bound recordable by **at-most**.

- *New assertions can be inferred about other individuals, usually ones related to b via roles.* For example, if an **all(friends,HAPPY)** description applies to some individual $x$, and now $y$ is asserted to be a **friends**-filler for $x$, then we can infer that $y$ is an instance of **HAPPY**.

To support the first task above, PROTODL might use the function **ConsistentW?**[7], which returns true to indicate that information to be added to an individual is consistent. For the second task, we use function **Recognizes?**. For the third and fourth tasks, which involve individuals other than the one on which the update has occured, we use function **InferFrom**.

These functions are used to implement the various ASSERT KBMS operations, as well as the **ClassifyIndividual** function, which is used by PROTODL to find the lowest classes in the hierarchy to which this individual belongs.

As in the case of concept reasoning, we use the **and** constructor to drive the processing, and we modularize the implementation so each kind of constructor **Q** has its own set of functions: $\boxed{\textbf{Q::Recognizes?}}$, $\boxed{\textbf{Q::ConsistentW?}}$, $\boxed{\textbf{Q::InferFrom}}$. It now becomes more important to distinguish constructors, like **all** and **at-most**, that have an associated single role, in contrast to generic constructors such as **one-of** (which has no roles associated) and **same-as** (which has many roles, none of which is special). For the former, so called *branch constructors*, the above functions will have arguments that include not just the individual and the normalized representation of **Q**-terms, but also the fillers and closed information about the role. (This is done to prevent repeated retrieval of these values by each such constructor.) Therefore, the general form of **and::Recognizes?**, provided in PROTODL, is

```
and::Recognizes?(b:<Individual>,NC:<NormalizedConcept>)returns <Bool>
    { For every Q(T) in getNonbranchComponents(NC) do
        if not Q::Recognizes?(b,Q(T)) then return false;

    For every role p in  getRestrictedRoles(NC) do
        {F:= get_fillers(b,p);
          clsd? := is_closed?(b,p);
          For every  Q(T) in getBranchComponents(NC,p)
            {if not Q::Recognizes?(b,Q(T),F,clsd?)
              then return false};};
        return true;  }
```

---

7. Later, when we introduce incremental reasoning, we will identify a number of variants of this function.

The functions **and::ConsistentW?** and **and::InferFrom** are similar in structure, and are not presented due to space limitations. It is important to note that the **ConsistentW?** functions return false only to indicate that they are unable to prove *conclusively* that the individual is consistent or inconsistent; if an inconsistency is found, an **InconsistentExcn** is raised. Also, after an update there may be several individuals about which incompatible information will be asserted. The system only guarantees to find *one* such case, and then rejects the update.

The following is an example of a specialized function for the case **Q=all**, which implements the first inference rule presented earlier

```
all::Recognizes?(b:<Individual>, all(r:<Role>,vr:<NormalizedConcept>),
            fillers:<SET(Individual)>, clsd?:<Bool>)  returns <Bool>
      {if not clsd? then return false   /*More fillers might come later*/
       else for every f in fillers do
               {if not and::Recognizes?(f,vr)
                then return false;}
       return true }
```

## 5.2 Reasoning with Incremental Updates

Since updates to individuals are incremental in nature, in order to improve efficiency we want to take into account the fact that the KBMS had already performed all the inferences up to, but not including, the current update. For example, if some individual **Tintin** has been asserted to be an instance of **all(pet,DOG)**, and already has pet-fillers **d1** and **d2**, then if the current update is ASSERT-FILLS(**Tintin,pet,Fido**), then we only need to propagate the information that pet-fillers are **DOGs** to **Fido**, since the others would have been processed earlier. And, if the current update is ASSERT-CLOSED(**Tintin,pet**), then no new role-filler information can be inferred from this because of the **all** restriction, so we should not even bother calling **all::InferFrom**. We will therefore distinguish the following three variants of **Q::InferFrom**:

**Q::InferFromFilling(<Individual>,<NormalizedQ-term>,<Role>,<Individual>)**
**Q::InferFromClosing(<Individual>,<NormalizedQ-term>,<Role>)**, and
**Q::InferFromAsserting(<Individual>,<NormalizedQ-Term>)**

plus similar variants of **Q::ConsistentW?**. For example

- **all:ConsistentWFilling?(ind, all(r,vr), r, newfiller)** invokes **and::ConsistentWAsserting?(newfiller,vr)** to check that the new filler does not contradict the role restriction **vr**;

- **all::ConsistentWClosing?(ind,all(r,vr),r)** just returns true, and hence is hopefully eliminated by the compiler;

- **all::ConsistentWAsserting?(ind,all(r,vr))** verifies that all r-fillers y of **ind** are consistent with **vr** by invoking **and::ConsistentWAsserting?(y,vr)**.

One might also consider variants of **Q::Recognizes?**, but, as we shall see, in our case these are of limited use since we do not propose to keep information on which parts of the recognition test had succeeded already, and therefore we need to start from the beginning.

## 5.3 Dependency Links

Clearly, the results of function calls, such as **Recognizes?**, on some individual **Bob** might change after an update to **Bob** itself. However, the membership of **Bob** in a class or **Bob**'s consistency may depend on facts asserted about *other* individuals in the database. For example, if all of **Bob**'s friends are known (i.e., the role **friends** is closed), and all but one of them was **MARRIED**, and now that holdout, **Larry**, is also asserted or inferred to be **MARRIED**, then **Bob** itself can be reclassified as an instance of **all(friends,MARRIED)**.

This means that without some special data structure, after every update of some individual an unsophisticated implementation has to reconsider *every* other individual in the KB.

One alternative, used in LOOM (MacGregor, 1986), is to keep track of all questions asked about an individual as part of the previous processing ("hits" and "misses"), and if answers to these do not change as a result of the update then no <u>re-processing</u> is needed. Another alternative is to use an elaborate truth-maintenance system (as available in KL-TWO (Vilain, 1985)), for each kind of judgment. In our opinion, both these approaches might however become very expensive in terms of space and often computation time, because they maintain too many details.

We follow an intermediate stance, first suggested by Peter Patel-Schneider for the CLASSIC implementation, which is intended to reduce needless re-testing while incurring relatively less "dependency maintenance" overhead. Essentially, we have a "coarse-grained" dependency structure, where one individual $e$ may point to another, $b$, if the result of a decision about the latter may change as a result of *some* (unspecified) additional information being added to the former. For example, in the presence of the definition **CanineLover≐all(pet,DOG)**, if **Tintin** has **Fido** as a **pet** filler, and neither **Recognizes?(Fido,DOG)** nor **not ConsistentW?(Fido,DOG)** return true, then we add a link of the form: **Fido** $\overline{\text{RecognizeDependsOnMe}}$ **Tintin**. Thereafter, any change of status of **Fido** will cause **Tintin**'s classification (with respect to **CanineLover**, as well as other pending classes) to be re-done. Similarly, we will have $\overline{\text{ConsistentDependsOnMe}}$ and $\overline{\text{InferDependsOnMe}}$ links between objects.

Note that when following such dependency links back, we do know that only properties of the fillers of some role might have changed, so that once again we could have variants **and::RecheckConsistentW?** and **and::RedoInferFrom**, which perform less work by omitting the checks involving constructors that are known not to be affected by such aspects (e.g., **at-most** only cares about the count of the fillers, not their properties).

In order to set dependency links, every function such as **Q::Recognizes?** needs to keep track of the list of individuals whose modification might change the result of the function. These values must eventually be appropriately linked with dependencies either by the functions themselves or by the calling environment.

## 5.4 Coordinating the Components

As should be clear from the above examples, the task of the top-level KBMS update operations (asserts) will include not just invoking the appropriate functions *on the individual being updated*, but also setting appropriate dependency links, gathering into queues other objects whose dependency links have been "tickled" by the latest update, and then repeating these operations on the other individuals brought into focus by inferences or dependency links.

An appropriate software architecture for representing this processing seems to be a blackboard model, where we have 5 lists to which functions can post tasks to be carried out:

- Two lists, `ToRecheckConsistencyList, ToRedoInferFromList`, for objects that are reached from individual $x$ via a corresponding kind of dependency link as a result of a change to $x$.

- A list, `ToPerformUpdateList`, which collects the additional inferences to be made that are found by the `InferFrom` and `RedoInferFrom` function calls. For simplicity, these are recorded as additional calls to versions of the three kinds of assert functions.

- A list, `ToReclassifyList`, which receives objects that need to be checked in case they can be classified further down the hierarchy; these objects come not just from dependency links but also from any additional asserts that had been triggered.

- Finally, a list `ToAddDependencyList`, that keeps track of dependency links that need to be added to the knowledge base; this list is augmented by the various functions, as mentioned in the previous section.

Because, for example, re-classification may cause new inferences, which may cause re-classification due to some other constructor, there is no linear order in which these lists can be processed. We may therefore have a loop where a "demon" removes an arbitrary object from a list, invokes the required function, and then repeat the process. The only requirement is that dependencies on `ToAddDependencyList` be also considered as having been installed, whenever chasing objects that may have been affected by an update. Note that, not surprisingly, it is hard to make any general statements about the termination of the above algorithm skeleton, since the various functions, especially `InferFrom`. are free to do whatever they want, including increasingly longer descriptions..

Several policies can help the performance of the system:

- In all cases, it is helpful to eliminate duplicates from the lists.

- Locality of context may improve performance; therefore grouping together operations to be performed on one object is advisable (e.g., gather together all the dependencies from object $y$ on `ToAddDependencyList`).

- If we expect inconsistencies to arise infrequently, then the `ToRecheckConsistencyList` can be processed at the end.

```
ASSERT-FILLS(ind, role, filler)
     {if  redundant information then signal RedundantExn;
       add_filler(ind,role,filler);  /* this may be rolled back by an error */
      initialize the blackboard lists to empty;

      ind-descr := Descriptor(ind);
       /* deal with consistency issues involving ind */
            and::ConsistentWFilling?(ind, ind-descr, role, filler);
              /* the preceding function will post dependencies if
                the answer is not a definite TRUE  */
            ToRecheckConsistencyList += getConsistentDependsOnMe(ind);
               /* this puts up for reconsideration objects
                         dependent on ind */
       /* check for inferences */
            and::InferFromFilling(ind,ind-descr,role,filler);
                /* this may post new updates and/or dependencies */
            ToRedoInferFromList += getInferDependsOnMe(ind)
       /* reclassify at least this individual */
            Reclassify(ind);
            ToReclassifyList += getRecognizeDependsOnMe(ind);
       /* Process tasks left on the blackboard */
            ProcessBlackBoard;
     }
       catch InconsistentExcn
            {roll back update; print error msg; }
```

Figure 2: PROTODL pseudo-code for ASSERT-FILLS.

Figure 2 contains the pseudo-code for ASSERT-FILLS, which is offered by PROTODL. Notice the advantage in the above program of using exception handling and propagation up the *dynamic* function call hierarchy as a way of dealing with inconsistency: wherever it is detected by `ConsistentW?`, the exception passes up through levels of function calls, all of which are interrupted, until a handler is encountered — usually by an external or internal ASSERT operation; this issues appropriate error messages, resets local changes made, and then re-raises/propagates the exception further up, until the top level is reached.

## 5.5 Extending Individual Reasoning – An Example

Let us consider the extensions needed for a rather complex new concept constructor: **same-as**. The semantics of **same-as**($[f_1, \cdots, f_n], [g_1, \cdots, g_m]$), as introduced earlier, is that it denotes individuals for which the two attribute chains $[f_1, \cdots, f_n]$ and $[g_1, \cdots, g_m]$ have the same known filler. (The attributes must each have exactly one filler.)

This constructor is very useful in representing the relationship between actions and sub-actions. For example, suppose we try to define the action of `BUY`ing in terms of two `GIVE`s:

we would assert that BUY is subsumed by **and(all(giving1,GIVE),all(giving2,GIVE))**, and then add further constraints about the attributes of BUY (such as buyer and seller) and those of GIVE (such as giver and receiver), including **same-as([buyer],[giving1 giver])** and **same-as( [seller],[giving1 receiver])**.

The procedures that need to be written to extend the implementation are:   **same-as::-Recognize, same-as::InferFromFilling, same-as::-InferFromClosing, same-as::-InferFromAsserting, same-as::ConsistentWFilling?, same-as::ConsistentWClosing?,** and   **same-as::ConsistentWAsserting?**. In addition, some previously written procedures may need to be augmented.

Rules of inference for $\longrightarrow$ and $\not\longrightarrow$ can be presented as a way of describing the tasks performed by the above functions. However, in this case, the functions also need to worry about dependency pointers, so the mapping is less direct.

The following inference rule describes the standard recognition procedure for individuals (there are additional rules for subsumption of course):

$$KB \vdash b \longrightarrow \mathbf{fills}([f_1, \cdots, f_n], e)$$
$$KB \vdash b \longrightarrow \mathbf{fills}([g_1, \cdots, g_m], e)$$
$$\overline{KB \vdash b \longrightarrow \mathbf{same\text{-}as}([f_1, \cdots, f_n], [g_1, \cdots, g_m])}$$

If the recognition function can only follow a chain up to an individual $e$, which is missing the filler of the next attribute in the chain, then a dependency from $e$ must be recorded.

```
same-as::Recognizes?(b, same-as(chain1,chain2))
  { (e1,p') :=  follow chain1 from b,  returning the furthest
        individual e1  reached, and the  remainder of the chain p'
         which was not possible to follow;
     if not(p'=nil)
        then {  post dependency e1 RecognizeDependsOnMe b; return false;}
     (e2,q') := follow chain2 from b,  as above;
     if not(q'=nil)
        then {  post dependency e2 RecognizeDependsOnMe b; return false;}
     if (e1=e2) & p'=q' then  return true else return false.
  }
```

This function will be called from inside **and::Recognizes?(b,C)** as follows:

```
   ...
for every same-as(chn1,chn2) in get_sameas(C)
   if not same-as::Subsumes?(same-as(chn1,chn2), Descriptor(b))
   then if not  same-as::Recognizes?(b,same-as(chn1,chn2))
       then return false.
    ...
```

Reasoning with **same-as** concerning individuals is actually more complex, because there are additional rule of inference, such as

$$KB \vdash b \longrightarrow \mathbf{fills}(chain1, e)$$
$$KB \vdash b \longrightarrow \mathbf{same\text{-}as}(chain1, chain2)$$
$$KB \vdash e \longrightarrow \mathbf{same\text{-}as}(chain3, chain4)$$
$$\overline{KB \vdash b \longrightarrow \mathbf{same\text{-}as}(chain1 \circ chain3, chain2 \circ chain4)}$$

Applying this rule straightforwardly to verify whether some $b \longrightarrow \textbf{same-as}([f_1, \cdots, f_n], [g_1, \cdots, g_m])$ requires one to try all possible ways of dividing up the chain $[f_1, \cdots, f_n]$ into subchains, and for each, one needs to try to derive the subchain equalities alternately from individual fillers along the chain, or from explicit assertions of **same-as** descriptions on individuals. In this cases, the implementor, in consultation with the knowledge engineer who is to use the extended system, would make a judgment on the necessity of implementing this more expensive inference. The rules of inference can be used to describe the deductions that have been implemented and those that have not.

In order to check consistency, we need to consider the three kinds of updates: assertions, fillers, and closings. When an equality is first asserted of an individual, we would invoke:

```
same-as:: ConsistentWAsserting?(ind, same-as(chain1,chain2))
   { (e1,p') := follow chain1 from ind;
      if not(p'=nil)
         then { post dependency e1 ConsistentDependsOnMe ind; return true;}
      (e2,q') := follow chain2 from ind;
      if not(q'=nil)
         then { post dependency e2 ConsistentDependsOnMe ind; return true;}
      if e1=e2 then return true else signal InconsistentExcn;
      }
```

Since the attributes can have at most one value, these attributes get closed automatically when a filler is provided, so there is usually no explicit closing of the role that can affect **same-as**, and there is no need for **same-as::ConsistentWClosing?**.

Finally, if `C` includes as a conjunct **same-as(chain1,chain2)**, then **and::ConsistentW-Filling?(ind,C,p,newfiller)** will have to invoke

```
same-as::ConsistentWFilling?(ind,same-as(chain1,chain2),p,newfiller)
    {if (member(p,chain1) OR member(p,chain2) )
       then same-as:ConsistentWAsserting?(ind,same-as(chain1,chain2))}
```

Equalities can lead to inferences when one of the chains is completely known, and all values but that of the last attribute on the second chain are known, as illustrated by the following example:

$$\{\, a : \textbf{same-as}([q]\,,\,[p,r]),\ \ (a,e):q,\ \ (a,b):p \,\} \ \models\ b \longrightarrow \textbf{fills}(r,e)$$

This inference requires us to set up $\overrightarrow{\text{InferDependsOnMe}}$ links to watch for cases when we have sufficient information to make the inference.

We point out that **same-as** presents one example of a situation where a performance penalty is paid for separating out consistency checking, recognition and inference: each of these functions attempts to traverse the individuals along the two chains of the equality in an effort to reach the ends. This price is negligible if there are relatively few individuals with **same-as** conditions attached, or if the chains are usually only one or two attributes long, as is the case for the application of **same-as** illustrated earlier. Otherwise, we can introduce caching to speed up the code: associate with every equality and individual $b$ two pairs $(e_1,r_1)$ and $(e_2,r_2)$ representing the ends $e_i$ reached on each chain, and the remainder of the chains $r_i$ left to be traversed. (So if the chain is completely known, $r = nil$.) A final alternative is to add the code for **same-as::InferFrom** into **same-as::ConsistentW?** function, so the latter performs all the deductions involving **same-as**. The disadvantage of this approach, as of many optimizations, is the loss of modularity.

Finally, we mention that other extensions of reasoning at the individual level that we have considered include "database aggregate functions" like *sum* (e.g., **sum([departments budget], totalBudget)** can be used to model that **totalBudget** is the sum of the values for **budget** fillers for all department fillers); and epistemic constructors that allow one to query the state of knowledge (Donini et al., 1998) (e.g., **known-all(friends, known-at-most(1,pets))** recognizes individuals all of whose *known* **friend** fillers have at most one pet recorded in the KB, without having to have the roles be closed).

## 6. Conclusions

We began from the hypothesis that no "perfect" DL will ever be built, because of the need for application-specific reasoning, and potential incompleteness of reasoning due to the expressiveness-(tractability/decidability) trade-off. We argued that some of these issues are best attacked on a per-application basis. To resolve this problem, we proposed the use of an *extensible* DL-KBMS, where one tries to go as far as possible with an initial set of well-understood concept constructors, and then, when encountering unsolvable expressiveness problems (Doyle & Patil, 1991), add new concept constructors to overcome them.

We have also pointed out the limitations of this approach, which include not being conducive to including new forms of reasoning such as abduction, contexts, etc., and having difficulties with complete inferences for useful concept constructors that require reasoning by contradiction, and are best handled in the alternative DL reasoning paradigm – tableaux.

### 6.1 Implementation Status

Prototype implementations of aspects of both concept and individual reasoning in PROTODL have been carried out at Rutgers. However, experience with all fielded systems indicates that there is an order of magnitude more work to be done in making a system usable by others than their developers. For this reason, our goal is to add the results of the PROTODL research to an existing DL-reasoner. In fact, several ideas have been transfered, with the collaboration of Charles Isbell, to the newest version of the CLASSIC system released by AT&T Research. In particular, CLASSIC supports *test-defined* concepts – ones which allow the recognition of individuals through the use of an arbitrary LISP function. (This function can be seen as the combination of the **Recognizes?** and **ConsistentW?** functions discussed in the present paper.) In the newest version of CLASSIC, one can simulate the addition of new concept constructors by using them after the keyword **test-c**. For example, the concept **all(vacation,dateRange(1996/6/1,1996/8/31))** would be represented as

    (test-concept dateRange vacation '( (1996 6 1) (1996 8 31)) )

Although not all aspects of the protodl system have been implemented, the current extensions (Borgida et al., 1996) allow significant subsumption reasoning to be done for test extensions, and thus provide classic with the bases for extensible reasoning.

## 6.2 Related Work

In order to incorporate new concept constructors into a reasoner, we need to extend the implementation. One approach to this would be to offer some form of "declarative description" of the inferences to be performed, and then have a meta-interpreter which executes them. In fact, such approaches have been tried in the past for other kinds of representation formalisms (Greiner & Lenat, 1980; Genesereth, 1983). Except for cycl (Lenat & Guha, 1990), which allows the addition of new forms of inference rule schemas in First Order Predicate Calculus, we see little evidence that such a meta-interpreter has a chance of being nearly as efficient as custom-built implementations, so we have opted for a different approach.

Joshua (Rowley, Shrobe, & Cassels, 1986) is also an effort at providing extensible reasoning, which allows the user (knowledge system engineer) the ability to change at compile-time the implementation of any or all of the elements of the *protocol of inference*, which describes the reasoning of the system. Joshua is close in spirit to our work in the sense that it tries to maintain a uniform "knowledge level" view of the system, and because it identifies, in the "protocol of inference", the specific aspects of the system that can be customized through (re)programming of Lisp functions. Our efforts differ from Joshua in that we are interested in DLs (Joshua's protocol was concerned mostly with rule triggering and truth maintenance), we wish to support incrementing the syntax of the knowledge-level interface, and we also care about the semantics of the extension.

Gaines has also advocated the utility of a declarative specification and of a clean, extensible modularization for a DL-reasoner (Gaines, 1993). At the concept level, one difference between our approaches seems to be that protodl only assumes that the concepts will form an upper-semilattice (the **and** constructor is built-in), so that a wider variety of inferences can be implemented for new constructors. At the individual-reasoning level, protodl starts from a much more restricted basis, and uses its extensibility to deal with such aspects as "propagations" (e.g., our example dealing with **same-as** reasoning appears to be built-in in KRSn). On the other hand, KRSn has built-in support for rules and exceptions to them, which are an important component for any knowledge-based environment.

Finally, it has been suggested that the tableau-based approach, such as that of crack, is essentially extensible through the addition of new "completion" rules (Bresciani et al., 1995), which are traditionally used to build a model of a certain knowledge base, or prove its inconsistency. These tableau completion rules can also implement incomplete reasoners by using only a subset of the inferences. As with concrete domain extensions (Section 3.5), the advantages of extensible tableau techniques lie in a clean formalism that can lead to complete reasoning, while the disadvantages involve language extensions that do not have negation (e.g., clasp – see Section 3.4), and, for the moment, the lack of experience with large A-boxes and incremental updates. One of the most exciting (though very likely difficult) future prospects is combining the two implementation paradigms, and their extensibility features.

### 6.3 Summary

We have advocated an approach to extensible DL reasoning and implementation, for the normalize-compare paradigm, which has two components: a declarative specification and a modularized implementation framework. The specification is offered using rules of inference in the "natural semantics" style, and a heuristic methodology suggests various categories of rules to be looked for. These rules often correlate well with the implementations of the various functions, but PROTODL offers the implementor the opportunity to use a very different implementation. The later is needed in the case of constructors whose argument is for example some regular expression – as in the case of plans or strings, when the implementation needs to use some kind of finite automaton representation, since regular expressions have no "normal form".

We have modularized the software architecture of PROTODL reasoner, so that for every new concept constructor added to the language, there is a well-defined set of functions that need to be implemented. These, in turn, sometime have detailed skeletons composed of other, smaller functions, which we have abstracted after analyzing a variety of implementations. The invocation of these functions is organized by the built-in functions of PROTODL, usually involving the constructor **and**. We have paid particular attention to supporting efficient implementations, by offering the implementor quite a lot of freedom within the confines of our major functions. For individual reasoning, this is the first paper to consider the need to be efficient in the face of incremental updates of DL-KBMSs, which live with the open-world assumption, and which use concept descriptions to infer new properties, rather than just verify the correctness of individual facts. Our solutions involved proposing function variants based on the form of the update, and the use of various kinds of associated dependency links.

The major open areas involve adding to this framework role constructors, epistemic rules (like those in CLASSIC, characterized by Donini et al (Donini et al., 1998)), the ability to express at least simple recursive declarations for primitive concepts (e.g., "the parents of a person are persons"), and connections with the tableau-based implementations for DL reasoning.

### Acknowledgements

I am very grateful to Ron Brachman, who, among others, joined me in the initial explorations of the PROTODL idea; to Daniel Kudenko, who implemented a significant part of the individual reasoning; to Charles Isbell, who implemented the extensibility features of CLASSIC 2.3; and to Peter Patel-Schneider, for years of discussion on the subtleties of the CLASSIC language and implementation. Extremely useful comments on the presentation and organization were provided by Peter Clark and the anonymous reviewers.

This work was supported in part by NSF grants IRI-91-19310 and IRI-9619979.

### Appendix A. A Generic Conjunction Function.

The following pseudo-code for `Q::Conjoin` shows how one can construct this function from several, smaller functions.

```
/* Conjoin the new term Q(T) onto the already-normalized description This*/

Q::Conjoin(T:<NormalizedQ-term>,This:<NormalizedConcept>){

    old := get_Q(This) /* find the part of This dealing with constructor Q*/
    if not(old=nil)
    then if  Q::SubsumesSame?(T,old)
        then signal RedundantExn;
     T := Q::ConjoinToSame(T,old); /*  combine T with old*/

    if  Q::SubsumesDifferent?(T,This) /* T is implied by other constructors */
    then signal RedundantExn;

    T:= Q::ConjoinToDifferent(T,This) /* obtain a possibly stronger T*/
    if (type(T) ≠ Q) /* the stronger description might use another constructor*/
        then {post T to ImplicationsToDoList; /* for later processing*/
              return; }
    Q::ConsistentWithDifferent?(T,This) /* Check if T is coherent with the rest
                                     of This;  if not, an exception is raised.*/

    put_Q(T,This) /* add the description to the normalized descriptor This*/
    Q::FindOtherImplications(T,This,ImplicationsToDoList)
            /*  add any additional constructors implied by T and This */
```

Figure 3: Pseudo-code of a generic conjunction function.

Before we consider the individual functions introduced above, we remark that if the constructor **Q** implies terms built with other constructors (i.e., it adds entries to the `ImplicationsToDoList`), then the **and**::Conjoin function needs to be augmented with a loop at the end, which conjoins to `ontoNC` all the normalized terms on the `ImplicationsToDoList`:

```
    while notEmpty(ImplicationsToDoList)
      { another := removeAnElement(ImplicationsToDoList);
         suppose type(another)=Q;
      Q::conjoin(another,ontoNC)
```

Note that this processing may itself add new entries to the `ImplicationsToDoList`. In the case that there are many additions being done to the ImplicationsToDoList, a useful optimization might be to test for and remove redundancies from ImplicationsToDoList. If there are very few constructors making additions, a different optimization might be applied: add the above code only to the end of the corresponding constructor's `Conjoin` program.

The following is the intended purpose of the component functions used in the **Q**::Conjoin function in Figure 3. Note that for any particular constructor (e.g., **all**), not every such function needs to be implemented. When using a good optimizing compiler, default programs for these could be provided, which return constant values. Otherwise, `Conjoin` itself should be edited.

- **Q::Universal?(T)**: *Is* T *equivalent to the top of the hierarchy for that set of values?* For example, **at-least::Universal?** returns true when processing **at-least(0,players)**. This is used to eliminate unnecessary constructs from the normal form.

- **Q::Incoherent?(T)**: *Is* T *incoherent by itself?* For example, **some(players ,NOTHING)** is incoherent since NOTHING has no instances.

- **Q::IncoherentWithDifferent(T,This)**: *Is there an inconsistency if we consider conjoining with other constructors?* For example, when **Q=at-least**, conjoining **at-least(3,players)** to a concept already containing **at-most(1,players)** would lead to inconsistency.

- **Q::SubsumesSame?(T,oldTs)**: *Is* T *implied by the* **Q**-*constructed terms already seen in oldTs?* This is basically the "structural subsumption" part. For example, if **Q=at-least**, and **oldTs** has **at-least(4,players)**, then **at-least(3,players)** is redundant

- **Q::SubsumesDifferent?(T,This)**: *Is* T *implied by other constructors?* For example, **at-least(1,players)** is implied by **some(players, OLD)**.

- **Q::ConjoinToSame(T,OldTs)**: *Merge* T *with any preceding terms built with* K. For example, when we combine **all(players,GENTLEMAN)** with **all(players,SCHOLAR)**, we obtain **all(players,and(GENTLEMAN, SCHOLAR))**.

- **Q::ConjoinToDifferent(T,This)**: *Produce a stronger* T *by using information from other constructors.* For example, if we are processing **some(player, TALL)**,then if the concept **This** already has **all(player,OLD)**, then **some::ConjoinToDifferent** returns the description of **some(players,and(TALL,OLD))**.

- **Q::FindOtherImplications(T,This,ImplicationsToDoList)**: *Add constructors which are implied because of* **Q***(T), possibly in conjunction with the rest of the concept,* **This**. For example, when adding **all(players,TALL)** to a concept with **at-least(3,players)**, **all::FindOtherImplications** would add **some(players,TALL)** to the list of constructors to be conjoined to **This**.